%% file: seo_456.tex
\documentclass[accepted]{uai2021} 

\usepackage[american]{babel}

\usepackage{times}
\usepackage{epsfig}
\usepackage{graphicx}
\usepackage{amsmath}
\usepackage{amssymb}

\usepackage[numbers, sort]{natbib}
\usepackage{commath}
\usepackage{amsthm, mathtools}
\usepackage{subcaption}
\usepackage{booktabs}
\usepackage{algorithm,algorithmic}
\usepackage{multicol}

\usepackage{siunitx,etoolbox}
\usepackage{tikz}
\usetikzlibrary{fit,calc} 
\newcommand*{\tikzmk}[1]{\tikz[remember picture,overlay] \node (#1) {};\ignorespaces}
\newcommand{\boxit}[1]{\tikz[remember picture,overlay]{
    \node[yshift=2pt,fill=#1,opacity=.10,fit={(A)(B)}] {};}\ignorespaces}
\newcommand{\hilight}[1]{\tikzmk{A}#1\tikzmk{B}\boxit{red}}

\usetikzlibrary{bayesnet,arrows.meta}
    
\makeatletter 
\newcommand{\subalign}[1]{%
  \vcenter{%
    \Let@ \restore@math@cr \default@tag
    \baselineskip\fontdimen10 \scriptfont\tw@
    \advance\baselineskip\fontdimen12 \scriptfont\tw@
    \lineskip\thr@@\fontdimen8 \scriptfont\thr@@
    \lineskiplimit\lineskip
    \ialign{\hfil$\m@th\scriptstyle##$&$\m@th\scriptstyle{}##$\hfil\crcr
      #1\crcr
    }%
  }%
}
\makeatother

\usepackage{hyperref}
\usepackage{cleveref}

\renewrobustcmd{\bfseries}{\fontseries{b}\selectfont}
\renewrobustcmd{\boldmath}{}
\newrobustcmd{\BF}{\bfseries}

\newcommand{\todo}[1]{\textcolor{red}{(\textbf{TODO}: #1)}}

\newcommand{\lyh}[1]{{\color{blue} (lyh: #1)}}
\newcommand{\mks}[1]{{\color{blue} (mks: #1)}}

\newcommand{\expnum}[2]{{#1}\mathrm{e}{-#2}}

\def\ie{\emph{i.e.}}
\def\eg{\emph{e.g.}}
\def\etal{\emph{et al.}}

\DeclarePairedDelimiterX{\infdivx}[2]{(}{)}{#1\;\delimsize|\delimsize|\;#2}
\newcommand{\kld}[2]{\ensuremath{D_{KL}\infdivx{#1}{#2}}}

\newcommand{\na}{\overline{\ba}}
\newcommand{\nw}{\overline{\bw}}

\input{def.tex}

\title{On The Distribution of Penultimate Activations of Classification Networks}

\makeatletter
\newcommand{\printfnsymbol}[1]{%
  \textsuperscript{\@fnsymbol{#1}}%
}
\makeatother

\author[1]{Minkyo Seo\thanks{The two authors equally contributed. This work was done while Minkyo Seo was visiting Kakao as a research intern.}}
\author[3]{Yoonho Lee\printfnsymbol{1}}
\author[2]{Suha Kwak}
\affil[1]{%
    Department of Computer Science and Engineering, POSTECH, Pohang, South Korea
}
\affil[2]{%
    Graduate School of AI, POSTECH, Pohang, South Korea
}
\affil[3]{%
    AITRICS, Seoul, South Korea
}

\begin{document}
\maketitle
\input{_0_abstract.tex}
\input{_1_introduction.tex}
\input{_2_analysis.tex}
\input{_3_related_works.tex}

\input{_4_applications.tex}
\input{_5_discussion.tex}

\bibliography{seo_456}
\end{document}

%% file: def.tex
\newcommand{\ba}{\mathbf{a}}

\newcommand{\bl}{\mathbf{l}}

\newcommand{\bw}{\mathbf{w}}
\newcommand{\bx}{\mathbf{x}}
\newcommand{\by}{\mathbf{y}}
\newcommand{\bz}{\mathbf{z}}

\newcommand{\bW}{\mathbf{W}}

\newcommand{\calL}{{\mathcal{L}}}

\newcommand{\Real}{\mathbb R}

\usepackage{color}
\definecolor{orange}{rgb}{1,0.3,0}
\definecolor{copper}{rgb}{1,.62,.40}
\definecolor{hotpink}{rgb}{1,0,0.5}
\definecolor{darkgreen1}{rgb}{0, .35, 0}
\definecolor{darkgreen}{rgb}{0, .6, 0}
\definecolor{darkred}{rgb}{.75,0,0}

\newcommand{\be}{\begin{eqnarray}}
\newcommand{\ee}{\end{eqnarray}}
\newcommand{\bee}{\begin{eqnarray*}}
\newcommand{\eee}{\end{eqnarray*}}

\newcommand{\matrixb}{\left[ \begin{array}}
\newcommand{\matrixe}{\end{array} \right]}

\newcommand{\argmax}{\operatornamewithlimits{\arg \max}}
\newcommand{\argmin}{\operatornamewithlimits{\arg \min}}

\newcommand{\E}{\mathbb E}

\newcommand{\defeq}{\stackrel{\mathrm{def}}{=}}

%% file: _0_abstract.tex
\begin{abstract}

This paper studies probability distributions of penultimate activations of classification networks.
We show that, when a classification network is trained with the cross-entropy loss, its final classification layer forms a \emph{Generative-Discriminative pair} with a generative classifier based on a specific distribution of penultimate activations.
More importantly, the distribution is parameterized by the weights of the final fully-connected layer, and can be considered as a generative model that synthesizes the penultimate activations without feeding input data.
We empirically demonstrate that this generative model enables stable knowledge distillation in the presence of domain shift, and can transfer knowledge from a classifier to variational autoencoders and generative adversarial networks for class-conditional image generation.
\end{abstract}

%% file: _1_introduction.tex

\section{Introduction}
Deep neural networks have achieved remarkable success in image classification~\citep{resnet,densenet}.
In most of these networks, an input image is first processed by multiple layers of neurons, whose final output, called \emph{penultimate activations}, is in turn fed to the last fully connected layer that conducts classification.
These networks are typically trained in an end-to-end manner by minimizing the \emph{cross-entropy loss}.
The penultimate activations are the deepest image representation of the networks and have proven to be useful for various purposes besides classification such as image retrieval~\citep{zhai_class_metric}, semantic segmentation~\citep{deconvnet}, and general image description of unseen classes~\citep{vggnet}.

This paper studies the penultimate activations of classification networks through a generative model. 
We derive this model by exploiting a dual relationship between the activations and the final classification layer weights, which results from the common practice of applying softmax to the output of a linear layer. 
Because of this, penultimate activations are determined entirely by their preceding layers, and they interact with the final classification layer in a predetermined way.
Using this fixed interaction, we can approximately recover information about penultimate activations using only the final layer's weights.



Because this generative model only uses final layer weights, it yields a compact representation of inter-class affinity with many potential applications.
It can serve as a lightweight knowledge transfer protocol, especially compared to previous frameworks requiring feeding data through networks.
Additionally, because our representation of class relations does not directly use data, it is more robust to domain shift and is potentially suitable for transferring knowledge in privacy-sensitive scenarios.
This representation's simple structure also enables transfer to learning modalities beyond supervised classification, such as serving as an effective data-driven prior for class-conditional generative models.
Furthermore, our model encodes information beyond the decision boundaries in the activation space, which has various potential applications, including anomaly detection and uncertainty estimation.

We experimentally demonstrate that our generative model of penultimate activations can be used for practical applications such as Knowledge Distillation (KD)~\citep{hinton_distilling_2015,ahn2019variational} and class-conditional image generation~\citep{kingma2014semi,davidson2018hyperspherical,miyato2018cgans}.
For KD, our model allows us to distill knowledge from a teacher network without feeding images forward through the teacher by generating its activations directly; this new approach to KD is complementary to the standard one~\citep{hinton_distilling_2015} and more robust against domain shift between teacher and student.
We also show that our model of penultimate activations in a trained classifier can be used as a data-dependent prior for a class-conditional image generation model, resulting in higher-quality synthetic images compared to those of vanilla models.

The remainder of this paper is organized as follows.
In~\cref{sec:analysis}, we analyze penultimate activations of classification networks and derive their probabilistic model.
After reviewing previous work related to our model of penultimate activations in~\cref{sec:related}, we apply our model to KD and conditional image generation in~\cref{sec:applications}.
We then conclude this paper with a discussion about limitations and future directions of our method in~\cref{sec:discussion}.

%% file: _2_analysis.tex

\section{Distributions of Penultimate Activations} 
\label{sec:analysis}

Consider a standard neural network that classifies data of $c$ different classes with ground-truth labels $i$.
We assume a balanced dataset where $p(i)= \frac{1}{c}$ for all classes $i=1, \ldots, c$.
We denote penultimate activations of the network by $\ba \in \Real^d$ and weights of the final fully-connected layer for classification by $\bW \in \Real^{d \times c}$;
this layer produces logits $\bW^\top \ba \in \Real^c$.
We denote columns of $\bW$ by $\bw_1, \ldots, \bw_c \in \Real^d$, and represent projections of $\ba, \bw_1, \cdots, \bw_c$ onto the unit hypersphere $\mathbb{S}^{d-1}$ using upper bars:
\begin{align}
\na \defeq \frac{\ba}{\norm{\ba}}, \hspace{10pt}
\nw_i \defeq \frac{\bw_i}{\norm{\bw_i}}.
\end{align}

\subsection{Analysis of Cross-Entropy Loss}
\label{subsec:xent}

We analyze how the common practice of minimizing cross-entropy loss affects the distribution of penultimate activations.
Our analysis reveals a close connection between the cross-entropy loss and a specific distribution of normalized penultimate activations, which can be described using only $\bW$, the last classification layer's parameters.

The column vectors $\bw_1, \ldots, \bw_c$ of $\bW$ can be interpreted as $c$ different prototypes, each of which represents a particular class. 
The network classifies a datapoint by comparing these prototypes against its penultimate activations.
This interpretation motivates us to derive a probability distribution of activations of class $i$ using $\nw_i$.
To this end, we first rewrite the cross-entropy loss in terms of penultimate activations:
\begin{align}
\label{eq:xent}
\calL_\textrm{xent} 
= -\log \frac{\exp(\norm{\bw_i} \norm{\ba} \nw_i^\top \na)}{\sum_j \exp(\norm{\bw_j} \norm{\ba} \nw_j^\top \na)}.
\end{align}

%

Eq.~\eqref{eq:xent} resembles Bayes' theorem:
Assuming $\norm{\ba}$ is a constant and $\norm{\bw}_i$ are the same for all $i$,
$\exp(\norm{\bw_i} \norm{\ba} \nw^\top \na)$ acts as an unnormalized joint probability for $(\na, i)$, 
and the denominator is the sum of all possible cases for class $j$.

The von Mises-Fisher (vMF) distribution, a well-known distribution in directional statistics, exactly takes this form of joint probability and is defined as
\begin{align}
\label{eq:vmf_definition}
\textrm{vMF}(\bx;\mu, \kappa) \defeq C(\kappa) \exp(\kappa \mu^\top \bx),
\end{align}
where $\mu \in \mathbb{S}^{d-1}$ is the mean direction, $\kappa \in [0, \infty)$ is a concentration term, and $C(\kappa)$ is a normalizing constant.
We write the cross-entropy in Eq.~\eqref{eq:xent} in terms of the vMF distributions:
\begin{align}
\label{eq:xent_to_logq}
\calL_\textrm{xent} = 
-\log \frac
    {\textrm{vMF}(\na; \nw_i, \norm{\bw_i} \norm{\ba})}
    {\sum_j \textrm{vMF}(\na; \nw_j, \norm{\bw_j} \norm{\ba})}.
\end{align}

\begin{figure} \centering
\begin{tikzpicture}
     \node[latent] (x) {$\bx$};%
     \node[latent, right=of x] (a) {$\na$};%
     \node[obs, above=of a, yshift=-0.3cm] (w) {$\bw$};%
     \node[obs, right=of a] (y) {$i$}; %
     \plate [inner sep=.3cm,xshift=.02cm,yshift=.1cm] {plate1} {(x) (a) (y)} {$N$}; %
     \edge {x} {a} 
     \edge {w, a} {y} 
    \draw[->, dashed, red] (w) to [bend right=30] (a);
    \draw[->, dashed, red] (y) to [bend left=30] (a);
\end{tikzpicture}
\caption{
    Plate notation representation of our model's structure.
    We assume that the network is trained to map $\bx$ to $i$, and exploit the structure of the last layer to approximately reconstruct activations using only weights $\bw$ and label $i$.
}
\label{fig:plate}
\end{figure}
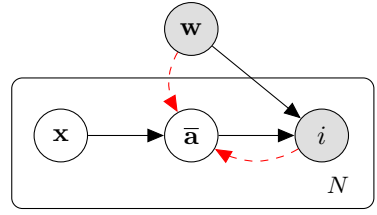

This motivates the following generative model which jointly models penultimate activations and labels:
\begin{align}
    q(\na, i) 
    = q(i) q(\na|i)
    = \frac{1}{c} \textrm{vMF}(\na; \nw_i, \norm{\bw_i} \norm{\ba}).
\end{align}
We see that the model $q(\na, i)$ forms a \emph{Generative-Discriminative pair}~\cite{ng2002discriminative} with the predictive distribution of the classification network:
\begin{align}
    \label{eq:gen_disc}
    &\argmax \log p(i|\na)
    = \argmin \calL_\textrm{xent} \nonumber \\
    &\approx \argmax \log \frac{q(\na, i)}{\sum_j q(\na, j)}
    = \argmax \log q(i |\na).
\end{align}
Eq.~\eqref{eq:gen_disc} shows that $q(\na, i)$ is closely related to the prediction of the classification network, and suggests that $q(\na|i)$ can be used as an approximation to the true posterior $p(\na|i)$.
Our modeling procedure is shown in \Cref{fig:plate}.
Note that while the network uses data $x$ during training, we do not assume access to data when approximately inferring activations $\na$.
By experiments in various domains, we will demonstrate that this simple model can transfer a substantial amount of information about the learned activation space.

For Eq.~\eqref{eq:gen_disc} to be an exact identity, the concentration parameter  $\norm{\bw_i} \norm{\ba}$ of each vMF component $q(\na|i)$ must be equal.
In the next section, we empirically verify to what extent this is true in trained classification networks.
In addition, to sample from $q(\na, i)$ using only the final layer parameters $\bW$, we treat $\norm{\ba}$ as a concentration hyperparameter and tune it using cross-validation on the downstream task. 


\subsection{Empirical Verification of Our Model}
In the previous section, we have suggested that the normalized activations $\na$ for each class follow the von Mises-Fisher distribution $q(\na|i)$.
We qualitatively verify this claim by visualizing penultimate activations of a classification network trained on the MNIST dataset~\citep{lecun-mnisthandwrittendigit-2010} and the vMF distributions derived from its final classification layer in~\Cref{fig:sunflower}.
The classification network consists of 4 convolution layers followed by the final fully connected layer that produces 2-dimensional penultimate activations (\ie, $\ba, \bw_1, \ldots, \bw_c \in \Real^2$).

\Cref{fig:sunflower} shows that in the early stages of training, normalized penultimate activations are not well aligned with vMF distributions.
However, as training progresses, they become grouped for each class and follow their corresponding vMF distributions.
This is in line with our analysis, in which we claimed that the normalized penultimate activations follow vMF distributions if the network is trained by minimizing the cross-entropy loss.

\subsection{Are Directional Statistics Sufficient for Classification?}
\label{subsec:sufficient}
Our approach to modeling normalized activations $q(\na|i)$ implicitly assumes that the directional vectors $\na$ and $\nw_i$ hold sufficient information for classification.

This assumption is empirically verified by quantifying how much the accuracy of trained classification networks drops when normalizing both $\ba$ and $\bw_i$.
To this end, we choose 9 networks pre-trained for the ImageNet classification task~\citep{imagenet} and measure their performance on the ImageNet validation set.
As summarized in~\Cref{tab:normalize_pretrained}, the performance drop by the normalization is marginal, especially when the network has more capacity.

We additionally provide an informal argument based on degrees of freedom for the sufficiency of directional statistics.
First, the norm of $\ba$ has no effect on the ranking of logits, thus not affecting the decision boundary in terms of logit ordering.
On the other hand, the norm of $\bw_i$ could change the decision boundary, but it has little influence on large networks.
Note that the direction vector $\nw$ accounts for $d-1$ of the $d$ degrees of freedom of the prototype vector $\bw \in \Real^d$.
Therefore, the fraction of the information that $\nw$ holds about $\bw$ roughly converges towards $1$ as $d$ increases ($\lim_{d \rightarrow \infty} \frac{d-1}{d} = 1$).
Since the dimension $d$ is typically very large in standard classification networks, we argue that the statistics of $\nw$ should hold sufficient information.

\input{figs/sunflower.tex}

\input{tables/normalize_pretrained.tex}

Furthermore, the sufficiency of directional statistics has been confirmed indirectly by the widespread use of hypersphere embedding in the face recognition literature~\citep{liu2019adaptiveface, fan2019spherereid} and the hardness metric based only on angles between activation and weight vectors~\cite{Chen20AVH}.
The fact that hypersphere embedding works in such domains supports our claim that classification can be done only with directional information.
Our analysis further suggests that even standard (unnormalized) networks may store most of their information in directional statistics.

%% file: figs/sunflower.tex

\begin{figure}
    \begin{subfigure}[t]{.3\linewidth}
    \centering\includegraphics[height=\linewidth]{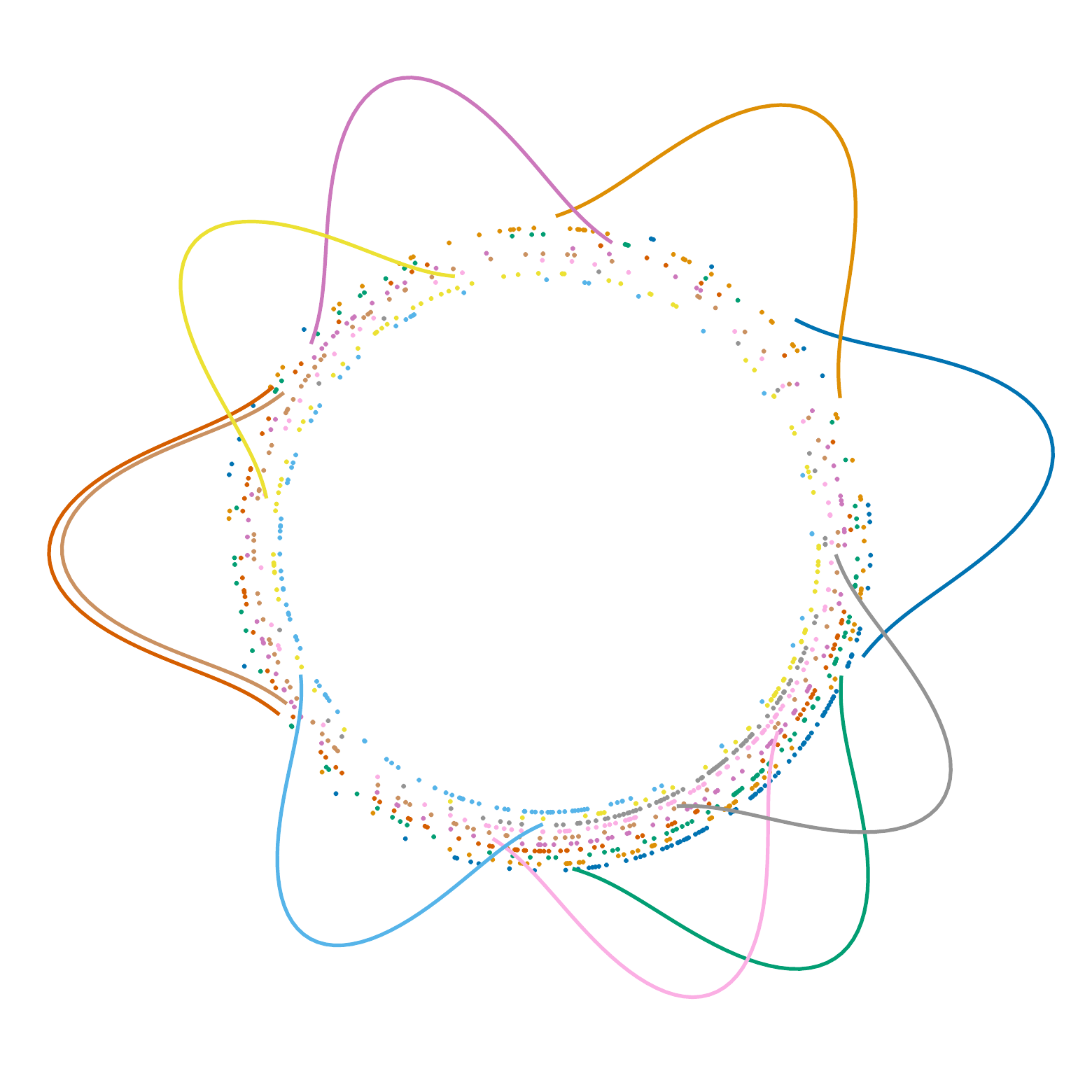}
    \caption{Epoch~1} \end{subfigure}
    \begin{subfigure}[t]{.3\linewidth}
    \centering\includegraphics[height=\linewidth]{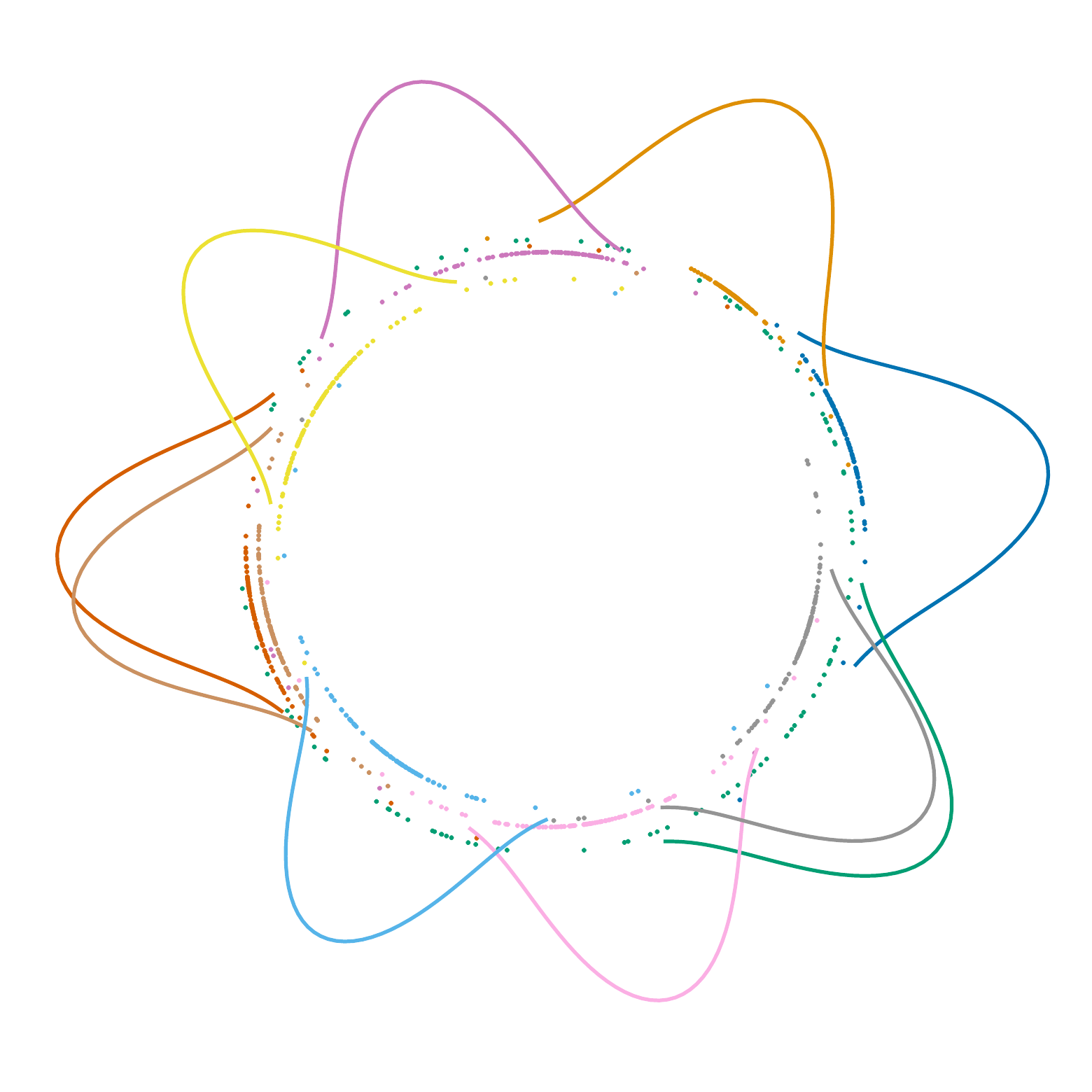}
    \caption{Epoch~2} \end{subfigure}
    \begin{subfigure}[t]{.3\linewidth}
    \centering\includegraphics[height=\linewidth]{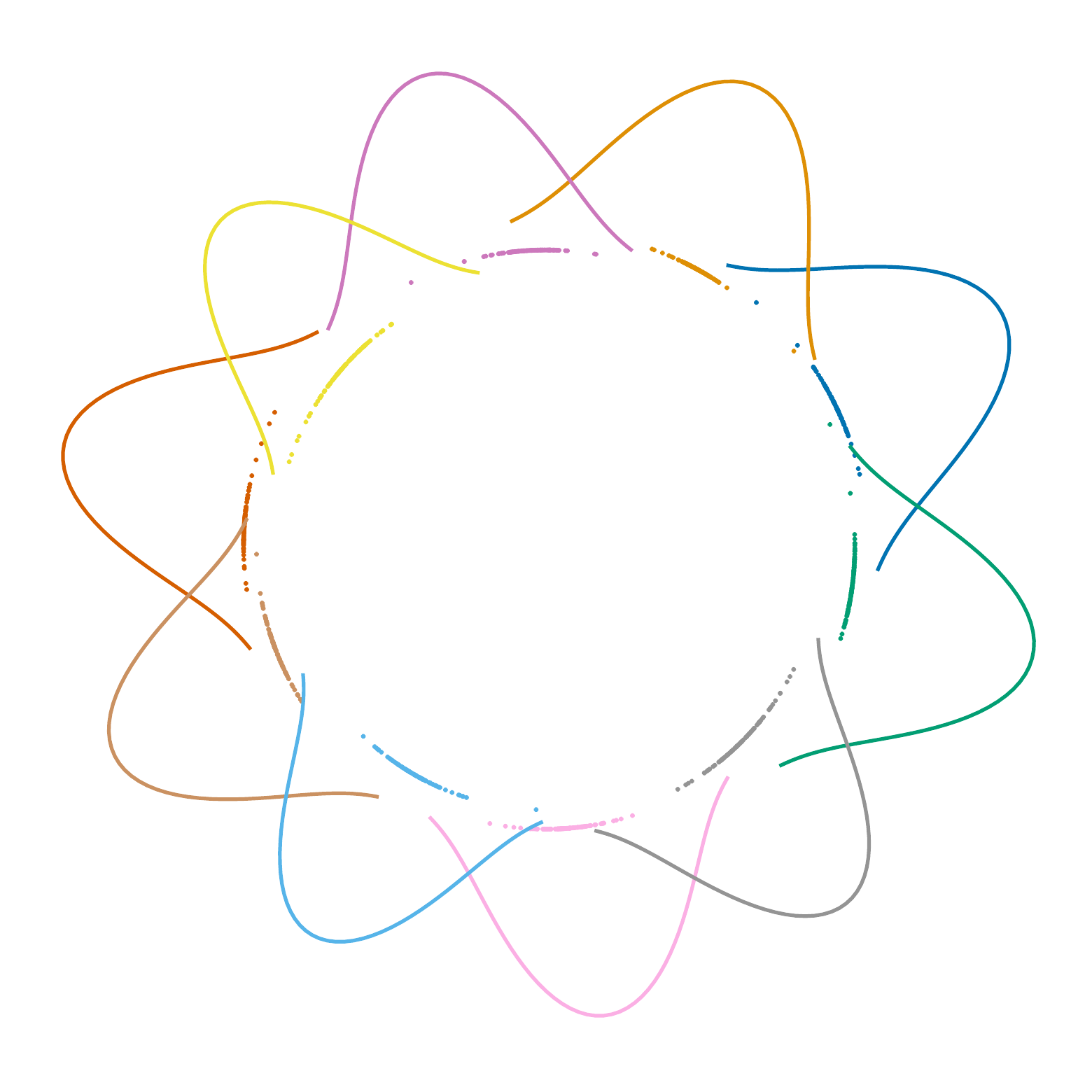}
    \caption{Epoch~30} \end{subfigure}
    \begin{subfigure}[t]{.07\linewidth}
    \centering\includegraphics[width=\linewidth]{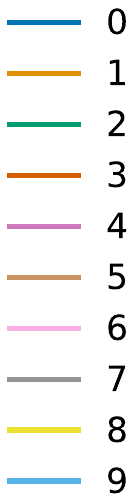} \end{subfigure}

    \caption{
        Visualization of penultimate activations and vMF distributions derived from a classification network trained on the MNIST dataset. 
        $\na$ of datapoints are represented by dots, and vMF distributions $q(\na|i)$ are drawn by solid lines.
    }
    \label{fig:sunflower}
\end{figure}

%% file: tables/normalize_pretrained.tex
\setlength{\intextsep}{0pt}

\begin{table*}
\caption{
Performance of various classification networks before and after normalizing $\ba$ and $\bw_i$ in top-1 accuracy on the ImageNet validation set. R: ResNet~\citep{resnet}, D: DenseNet~\citep{densenet}, S: ShuffleNet~\citep{shufflenet}, RX: ResNeXt~\citep{resnext}. 
}
\label{tab:normalize_pretrained}
\centering \begin{tabular}{lccccccccc}
\toprule
    & R-18 & R-50 & R-101 & R-152 & D-121 & D-201 & S-v2 & RX-50 & RX-101 \\
\midrule
Original   
    & 69.8 & 76.2 & 77.4 & 78.3 & 74.7 & 77.2 & 69.4 & 77.6 & 79.3 \\
Normalized 
    & 67.1 & 74.7 & 76.2 & 77.5 & 72.5 & 75.5 & 68.4 & 76.9 & 78.9 \\
\midrule
Drop rate 
    & -3.9\% & -1.9\% & -1.5\% & -1.1\% & -2.9\% & -2.2\% & -1.5\% & -0.9\% & -0.6\% \\
\bottomrule
\end{tabular} 
\end{table*}

%% file: _3_related_works.tex
\section{Related Work}
\label{sec:related}

\subsection{Understanding Deep Neural Networks}
\label{subsec:understanding}
Understanding what neural networks learn about data has been a fundamental problem in deep learning.
Previous methods have analyzed classification networks by optimizing an image to maximally activate a specific neuron~\citep{erhan2009visualizing,yosinski2015understanding} or maximize the predicted probability of a specific class~\citep{simonyan2013deep}.
A similar technique has been used to visualize the entire feature map of an image~\cite{Mahendran_2015_CVPR}.
For the same purpose, areas that most contribute to classification are estimated for each class through a weighted average of each activation channel~\cite{zhou2016learning}.
Our work also belongs to this line of research in that our model explicitly exhibits the learned class relations in a classification network.

While these previous methods offer high-level insights into the characteristics of deep neural network classifiers, they do not provide a way of using their insights to facilitate the training of other models.
Meanwhile, Yin~\etal~\cite{yin2020dreaming} propose to synthesize class-conditional images using knowledge of a trained classifier and exploits the generated images for knowledge distillation.
In contrast, our generative model can synthesize class-conditional \textit{activations} without such a costly image synthesis procedure.
We experimentally show that our method can extract the relationship between classes while being stable under distribution shift and transfer to different modalities such as generative modeling.

\subsection{Generative-Discriminative Pairs}
Generative classifiers model the joint probability $p(x, y)$ while discriminative ones model the conditional probability $p(y | x)$.
If two models belong to the same parametric family but respectively use the generative and discriminative criteria, the two are said to form a Generative-Discriminative pair~\cite{rubinstein1997discriminative,ng2002discriminative}.
In this context,
the work by Lee~\etal~\cite{lee2018simple} is the most similar to ours conceptually. 
They propose a generative model of activations that forms a generative-discriminative pair with a given classifier,
and apply the model to detecting out-of-distribution samples and adversarial attacks.
We also derive a generative model that approximately forms a generative-discriminative pair with any classification networks,
yet apply our model to transferring learned class relations to other networks and tasks.

\subsection{Learning with vMF Distributions}
As the vMF distribution is one of the simplest distributions for directional data, mixtures of vMFs have been widely used for clustering directional data \citep{Banerjee:2005:CUH:1046920.1088718,gopal2014mises}. 
For Bayesian inference of neural network weights, 
vMF distributions are used to model the directional statistics of the weights that are decomposed into radial and directional components~\cite{oh2019radial}.
Also, vMF embedding spaces have been studied for deep metric learning~\cite{hasnat2017mises} since such hypersphere embedding spaces are more desirable than conventional Euclidean spaces when their dimension is large.
Kumar~\etal~\cite{kumar2018von} used vMF distributions to reduce the large computations involved in normalizing the softmax for a set of words.

Our use of directional statistics differs from these previous methods: we use it as a tool for explaining the behavior of standard classification models rather than for specialized purposes like building a compact embedding space and computation reduction.

%% file: _4_applications.tex
\section{Applications}
\label{sec:applications}

This section demonstrates that our generative model of penultimate activations can be applied to two practical applications, KD~\citep{hinton_distilling_2015,ahn2019variational,romero_fitnets:_2014} and class-conditional image generation~\citep{davidson2018hyperspherical, odena2017conditional}.

\subsection{Class-wise Knowledge Distillation}
\label{app:kd}

This section describes how the generative model of activations can be used to develop a new algorithm for KD, and validates its effectiveness.

\subsubsection{Algorithm Details}
KD is the task of distilling knowledge from a teacher network $T$ to a student network $S$~\citep{hinton_distilling_2015}. 
Unlike most of the existing methods, our model enables KD without feeding data forward through $T$ by directly generating activations of a certain class. 
In detail, our model is used to approximate the average prediction of $T$ per class, which is represented as the probability of $T$'s prediction $y$ given class $i$ and estimated by
\begin{align}
p_T(y|i) 
= \int p_T(\na|i) p_T(y|\na) \,{d\na}
\approx \frac{1}{N} \sum_{j=1}^N p_T(y|\na_j),
\label{eq:mc_pyi}
\end{align}
where we employ Monte Carlo integration since the exact integral is intractable.
Also, each $\na_j$ is an \emph{i.i.d.} sample~from $\textrm{vMF}(\nw_i, \kappa)$, 
where $\kappa$ is set to 80 for all experiments by inspecting the empirical norm of the feature distribution on a teacher model.

The estimated $p_T(y|i)$ in Eq.~\eqref{eq:mc_pyi} quantifies the relationship between two classes $y$ and $i$ that is captured by $T$, and is employed as a target for KD in our method.
Recall that for teacher network $p_T$ and student network $p_S$, the standard KD loss \citep{hinton_distilling_2015} is
\begin{align}
\mathcal{L}_\textrm{KD} = -\E_{
\subalign{i, \bx  &\sim  p(i, \bx) \\ \by  &\sim p_T(y|\bx)}} 
    \left[ \log p_S(y|\bx) \right],
\label{eq:kd_loss}
\end{align}
where $y$ denotes prediction and $\bx$ and $i$ are data and label, respectively.
This loss is designed to minimize the KL divergence between $p_T(y|\bx)$ and $p_S(y|\bx)$ for each data $\bx$.
Unlike this data-wise KD, our approach is a Class-wise KD (CKD) whose objective is 
\begin{align}
\mathcal{L}_\textrm{CKD} = -\E_{
\subalign{i, \bx  &\sim  p(i, \bx) \\ \by  &\sim p_T(y|i)}} 
    \left[ \log p_S(y|\bx) \right],
\label{eq:new_loss}
\end{align}
where the categorical distribution $p_T(y|i)$ is given by Eq.~\eqref{eq:mc_pyi}.
Note again that while the standard KD objective in Eq.~\eqref{eq:kd_loss} requires a forward pass through the teacher network $T$ to compute $p_T(y|\bx)$, 
ours in Eq.~\eqref{eq:new_loss} utilizes the pre-computed distribution $p_T(y|i)$ without exploiting $T$ during training of $S$.
This property of CKD is useful especially when it is hard to conduct forward propagation through $T$ (\eg, online learning of $S$ with limited memory and computation power) or if there is domain shift between training datasets for $T$ and $S$ as demonstrated by experiments in \cref{subsubsec:kd_domain_shift}.
The overall procedures of the standard KD and our CKD are described in Algorithm~1 and 2, respectively, 
where the main differences between them are colored in red.

\subsubsection{Qualitative Analysis on the Effect of CKD}
\begin{figure}[t! ] \centering 
\begin{subfigure}[t]{.29\linewidth} \centering
    \includegraphics[width=\linewidth]{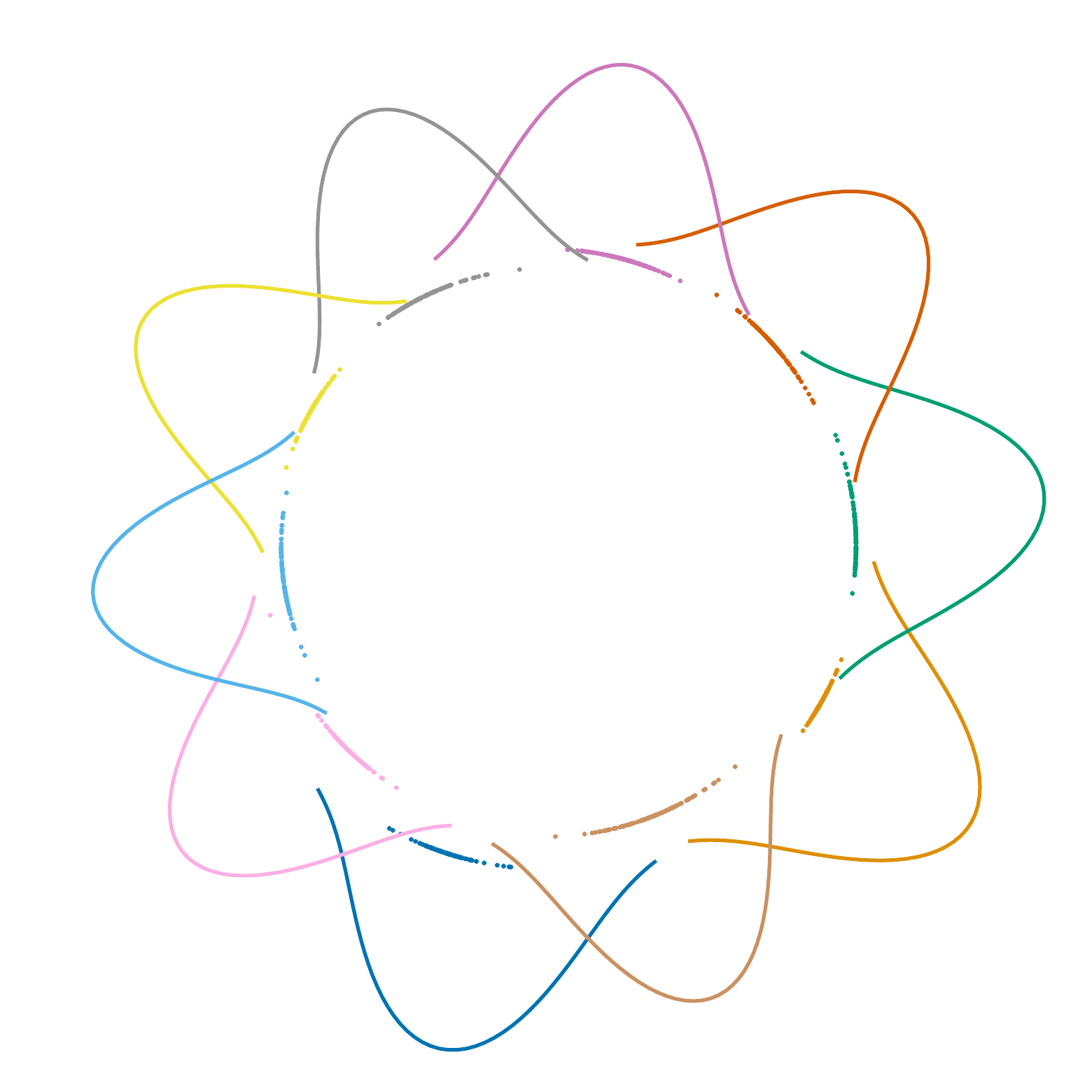}
\caption{Teacher} \end{subfigure} 
\begin{subfigure}[t]{.29\linewidth} \centering
    \includegraphics[width=\linewidth]{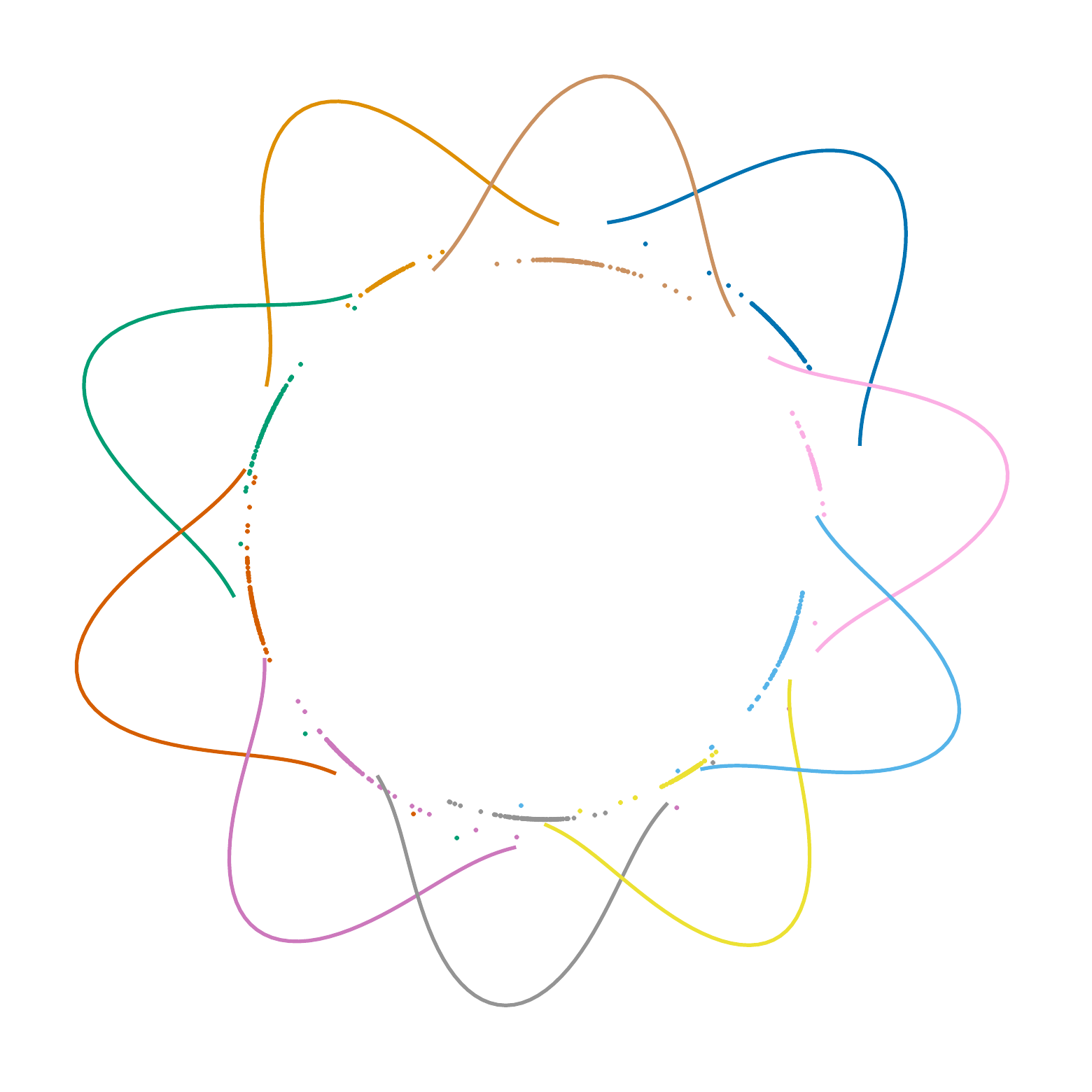}
\caption{Student (KD)} \end{subfigure}
\begin{subfigure}[t]{.29\linewidth} \centering
    \includegraphics[width=\linewidth]{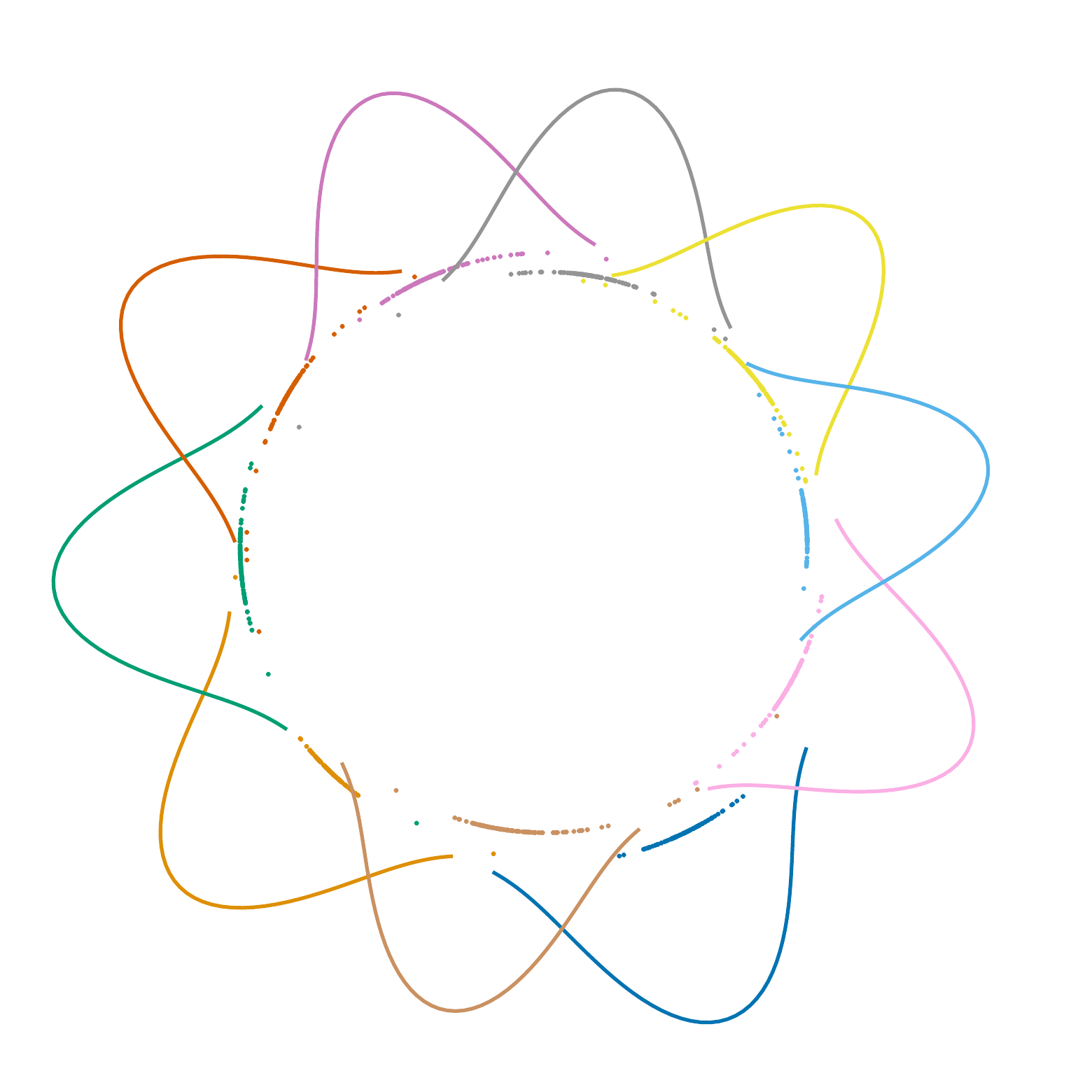}
\caption{Student (CKD)} \end{subfigure}
\includegraphics[height=0.29\linewidth]{figs/sunflower/lgnd_only.pdf}
\caption{
    Visualization of penultimate activations and their vMF distributions of teacher and student networks on the MNIST dataset.
    $\na$ of datapoints are represented by dots and vMF distributions $q(\na|i)$ are drawn by solid lines. 
}
\label{fig:sunflower_kd}
\end{figure}

To investigate which kind of information of $T$ is transferred to $S$ through KD or CKD, we qualitatively examine penultimate activations and their generative models of the three networks on the MNIST dataset. 
In this experiment, $T$ consists of 4 convolution layers followed by the final fully connected layer and produces 2-dimensional penultimate activations.
$S$ has the same architecture but with half the number of convolution kernels. 
From the visualization results in~\Cref{fig:sunflower_kd}, 
we observe that $T$ and $S$ have the same cyclic order of classes in the space of their penultimate activations.
This demonstrates that Eq.~\eqref{eq:new_loss} encourages $S$ to follow the inter-class relationships captured by $T$.

\noindent%

\begin{algorithm}[t] \begin{algorithmic}[1]
\caption{Knowledge Distillation \citep{hinton_distilling_2015}}
\label{alg:kd}
    \REQUIRE teacher network $\bx \mapsto p_T(y|\bx)$
    \REQUIRE student network $\bx \mapsto p_S(y|\bx)$
    \WHILE{not converged}
        \STATE $\bx, i \sim p(\bx, i)$
        \STATE \hilight{$p_T \Leftarrow p_T(y|\bx)$}
        \STATE $p_S \Leftarrow p_S(y|\bx)$
        \STATE $\mathcal{L}_\textrm{KD} \Leftarrow - p_T \cdot \log p_S$
    \ENDWHILE
\end{algorithmic} \end{algorithm}
\begin{algorithm}[t] \begin{algorithmic}[1]
\caption{Class-wise Knowledge Distillation (ours)}
\label{alg:ckd}
    \REQUIRE teacher network $\bx \mapsto p_T(y|\bx)$
    \REQUIRE student network $\bx \mapsto p_S(y|\bx)$
    \STATE \hilight{
        $p_T(y|i) \Leftarrow \frac{1}{N} \sum_{j=1}^N p_T(y|\na_j)$}
    \WHILE{not converged}
        \STATE $\bx, i \sim p(\bx, i)$
        \STATE \hilight{$p_T \Leftarrow p_T(y|i)$}
        \STATE $p_S \Leftarrow p_S(y|\bx)$
        \STATE $\mathcal{L}_\textrm{CKD} \Leftarrow - p_T \cdot \log p_S$
    \ENDWHILE
\end{algorithmic} 
\end{algorithm}

\subsubsection{Network Compression through KD}
\label{subsec:net_compress}
The effectiveness of CKD is first evaluated on the CIFAR-100 dataset~\citep{krizhevsky2009learning} in the scenario of network compression.
We adopt WRN-40-2 as $T$ and WRN-16-2 as $S$, both of which are introduced by~\cite{zagoruyko2016wide}, and follow the experimental protocol for network compression proposed in~\cite{tian2019contrastive}.
Follwing~\cite{hinton_distilling_2015}, we set the temperature for the KD loss to $4$ for all experiments.
The results of CKD are quantified and compared to other distillation methods in~\Cref{tab:KD}.
CKD outperforms most previous methods such as RKD~\citep{park2019relational}, SP~\citep{tung2019similarity}, and VID~\citep{ahn2019variational}.
These results demonstrate that CKD is capable of extracting useful knowledge from $T$.
In addition, CKD and the standard KD~\citep{hinton_distilling_2015} are complementary to each other and the performance is further enhanced by integrating them. 

\input{tables/KD.tex}
\input{tables/KD_resolution.tex} 

\subsubsection{KD in the Presence of Domain Shift} 
\label{subsubsec:kd_domain_shift}
Most KD techniques assume that $T$ and $S$ are trained with the same dataset or, at least, on the same domain. 
However, this assumption does not always hold in real-world settings, \eg, 
when the dataset used to train $T$ is not available due to privacy issues or when we train $S$ using streaming data that may be corrupted by various noises.
In those cases, the quality of knowledge extracted from $T$ in a data-wise manner may be degraded since $T$ assumes a data distribution different from what $S$ observes.

We argue that our CKD is more robust against such a domain shift issue since it performs KD without taking input data explicitly. 
We evaluate CKD and compare it to the standard KD~\citep{hinton_distilling_2015} on the CIFAR-100 dataset~\citep{krizhevsky2009learning} while simulating domain shift.
Specifically, we consider two different types of domain shift: photometric transform and downsampling. 
For the photometric transform, we randomly alter brightness, contrast, and saturation of input image with five different degrees $\in \{0, 0.2, 0.4, 0.6, 0.8\}$ of alteration, where $0$ means we use the original image without noise.
Also, for image downsampling, we reduce input image resolution with three different rates ($\times 0.75, \times 0.5, \times 0.25$) using nearest-neighbor interpolation.
$T$ is trained on the original dataset while $S$ is trained on its domain-shifted versions.
We use the same architectures as in \cref{subsec:net_compress}, using WRN-40-2 as $T$ and WRN-16-2 as $S$.

Experimental results are summarized in \Cref{tab:KD_domain}.
CKD consistently enhances the performance of the baseline using only ground-truth labels (``Label''). 
On the other hand, the standard KD (``KD'') deteriorates when the domain shift is significant.
We believe this result is mainly because the standard KD strongly depends on the data distribution.
On the other hand, the knowledge captured by CKD is still useful in the presence of domain shift since it extracts inter-class relationships directly from the weights of the final classification layer rather than relying on the data.



\subsection{Conditional Image Generation with Hyperspherical VAE}
\label{subsec:HVAE}

We use our generative model of penultimate activations to enhance class-conditional generative models.
Such models generate data $\bx$ using a latent variable $\bz$ together with a class label $i$.
While previous methods typically use the concatenated vector $[\bz; i]$ as an input to the decoder network,
we propose to instead use our learned model of activations as the distribution of $\bz$ given the label: $p(\bz|i)=q(\na|i)$.
In this section, this idea is applied to Hyperspherical Variational Auto-Encoder (HVAE)~\cite{davidson2018hyperspherical}, a latent variable model which performs inference using a vMF latent distribution.

This section first describes HVAE and its variant for conditional image generation, then illustrates how our model of activations is integrated with HVAE and improves the quality of generated images. 
The efficacy of our method is demonstrated on the MNIST dataset.

\paragraph{Baseline 1: Hyperspherical VAE (HVAE).}
HVAE is a latent variable model, which first computes the latent variable $\bz \in \mathbb{S}^d$ of a given datapoint $\bx$ using an encoder $q(\bz|\bx)$, then reconstructs $\bx$ from $\bz$ by a stochastic decoder $p(\bx|\bz)$.
HVAE assumes that $p(\bz)$ is a uniform distribution on the unit hypersphere $\mathbb{S}^d$. 
Accordingly, the encoder of HVAE is trained by maximizing the following lower bound of the evidence, usually called the ELBO:
\begin{align}
\E_{q(\bz|\bx)}[\log p(\bx|\bz)] - \kld{q(\bz|\bx)}{p(\bz)}.
\label{eq:hvae}
\end{align}

\paragraph{Baseline 2: HVAE Conditioned by Concatenation (HVAE-C).}
A straightforward way to extend HVAE to take class label $i$ into account is to concatenate $i$ to the end of the latent vector $\bz$.
We call this conditioned HVAE model HVAE-C.
Whereas HVAE assumes that $\bx$ is generated from $\bz$ alone, \ie, $p(\bx, \bz) = p(\bz)p(\bx|\bz)$, \mbox{HVAE-C} assumes that $\bx$ is generated from both $\bz$ and $i$, \ie, \mbox{$p(\bx, \bz, i) = p(i)p(\bz)p(\bx|\bz, i)$}.
We train HVAE-C by maximizing the following lower bound of the evidence considering $i$:
\begin{align}
\E_{q(\bz|\bx, i)} \left[ \log p(\bx|i, \bz) + \log p(i) \right] - \kld{q(\bz|\bx)}{p(\bz)}.
\label{eq:cvae}
\end{align}

\paragraph{Ours: HVAE Conditioned by Learned Prior (HVAE-L).}
Recall from \cref{subsec:xent} that one can utilize the weights of the last fully connected layer of a classification network to model a distribution of penultimate activations for a specific class $i$.
We employ this activation distribution conditioned on class $i$ as a learned prior for $\bz$ of HVAE.
This method is similar to HVAE-C in that the class information is involved in the process of generating $\bx$, but the two models differ in the way to integrate the information.
Unlike HVAE-C, HVAE-L generates $\bx$ from $\bz$ alone, yet the distribution of $\bz$ is determined by class label $i$, \ie, $p(\bx, \bz, i) = p(i)p(\bz|i)p(\bx|\bz)$.
Accordingly, it is trained by optimizing the following objective:
\begin{align}
\E_{q(\bz|\bx, i)} \left[ \log p(\bx|\bz) + \log p(i) \right] - \kld{q(\bz|\bx)}{p(\bz|i)}.
\end{align}
Also, the above objective differs from that of Eq.~\eqref{eq:hvae} since the two models assume different generation procedures.

\input{tables/scvae.tex}
\begin{figure}[t] \centering
\begin{subfigure}[b]{0.23\textwidth} \centering
    \includegraphics[width=\textwidth]{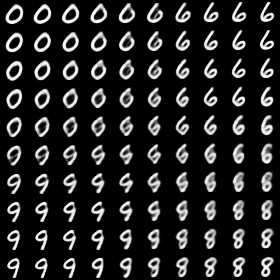}
\end{subfigure} \hfill
\begin{subfigure}[b]{0.23\textwidth} \centering
    \includegraphics[width=\textwidth]{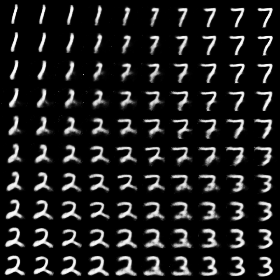}
\end{subfigure} \\
\vspace{8pt}
\begin{subfigure}[b]{0.23\textwidth} \centering
    \includegraphics[width=\textwidth]{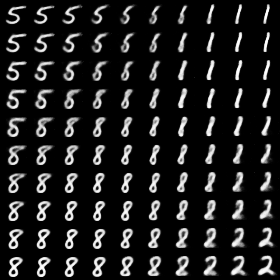}
\end{subfigure} \hfill
\begin{subfigure}[b]{0.23\textwidth} \centering
    \includegraphics[width=\textwidth]{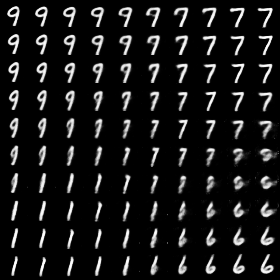}
\end{subfigure}
\caption{
    Visualization of the latent space of HVAE-L by interpolating between centers of different classes.
}
\label{fig:hvae}
\end{figure}

\begin{figure}[t]
\centering
    \includegraphics[width=0.99\columnwidth]{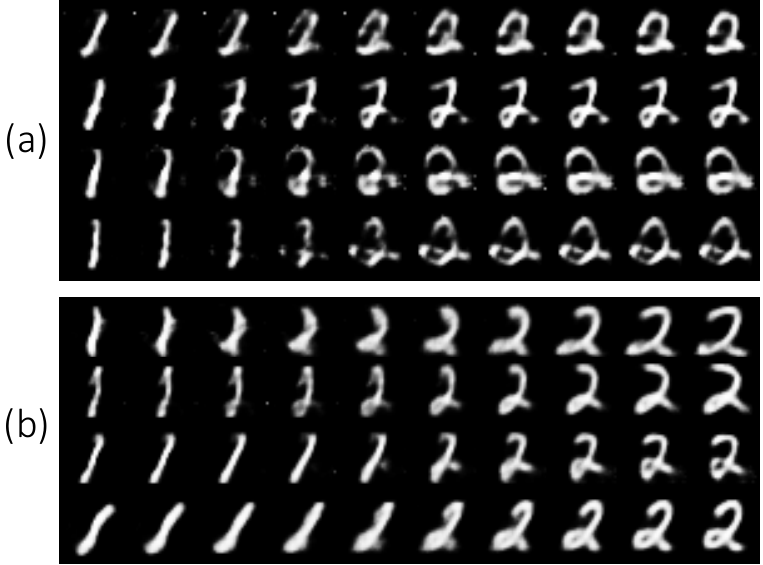}
\caption{
    Visualization of the latent spaces. (a) HVAE-C. (b) HVAE-L.
    For HVAE-C, input latent vectors are sampled through interpolation between class codes of $1$ and $2$ while fixing $\bz$.
    On the other hand, input latent vectors for HVAE-L are computed by interpolation between two points sampled from $p(\bz|i=1)$ and $p(\bz|i=2)$, respectively.
}
\label{fig:hvae_inter2}
\end{figure}

We compare our model (HVAE-L) against the two baselines (HVAE and HVAE-C) on the MNIST image generation task. 
Our experimental setup, including network architecture and hyperparameters, follows that of~\cite{davidson2018hyperspherical}. 
Specifically, both of the encoder and decoder of these models consist of three fully connected layers, whose output dimensions are $768-256-128-\text{dimension of } \bz-128-256-768$.
In addition, we ensure that the dimensionality of the latent vector $\bz$ is the same for all the models.
The prior distribution $p(\bz|i)$ of HVAE-L is derived from an MNIST classification network, whose architecture is the same with that of the encoder; the concentration parameter $\kappa$ of the prior distribution is set to $20$.
All the networks including the classification network are optimized by the Adam optimizer~\cite{kingma2014adam} with a learning rate of $\expnum{1}{3}$ and mini-batches of $64$ images. 

The performance of our model is summarized and compared with that of the two baselines in~\Cref{tab:VAE}, where HVAE-L outperforms both baselines. 
This result demonstrates that $q(\na|i)$, the vMF distributions of class-conditional activations can serve as a useful prior for class-conditional image generation of HVAE.
Specifically, we conjecture that the improvement by HVAE-L arises from the following properties of the prior.
First, the prior is derived from a classification network trained using examples of all classes, thus is aware of the affinity between different classes as well as variations within each class.
Second, the prior represents class identity and appearance variation jointly within a single latent space.
These two properties allow {{HVAE-L}} to exploit the latent space more flexibly and effectively.
In contrast, HVAE-C cannot take these advantages since it treats class labels as independent symbols and disentangles them from appearance variations.
The advantage of our prior model is also demonstrated qualitatively in~\Cref{fig:hvae} and~\Cref{fig:hvae_inter2}, where images generated by HVAE-L are more smoothly and naturally interpolated between different classes in the latent space than those of HVAE-C.


\begin{figure*}[!t] \centering
    \includegraphics[width=0.9\textwidth]{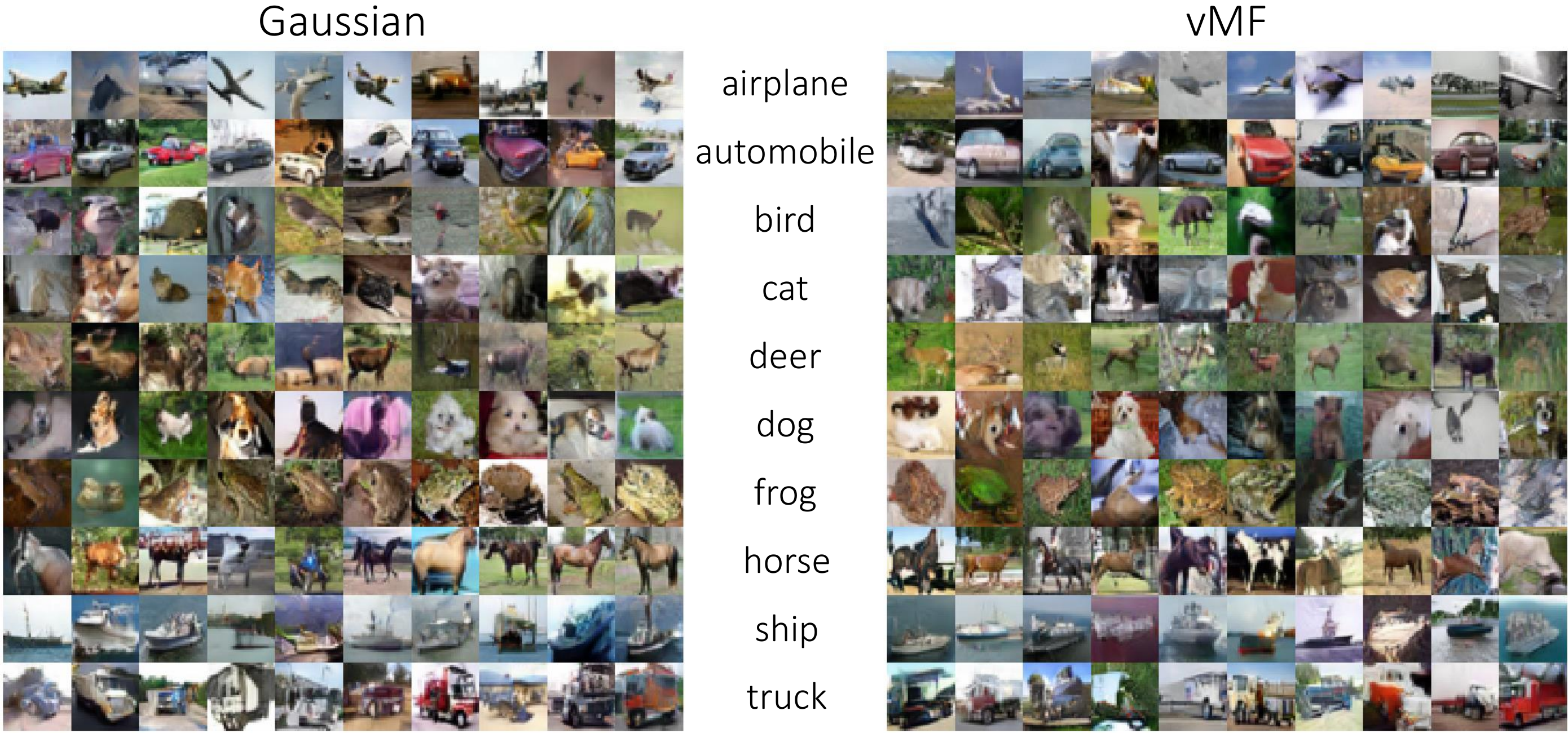}
\caption{Qualitative conditional image generation results using GAN variants on the CIFAR10 dataset.}
\vspace{0.3cm}
\label{fig:gan_img}
\end{figure*}


\subsection{Conditional Image Generation with GANs}
In this section, our generative model of penultimate activations is utilized as a class-conditional prior for conditional Generative Adversarial Networks (cGANs)~\cite{miyato2018cgans}.
Most cGANs independently sample a class label $i$ and a latent vector indicating a specific appearance of the class ~\citep{miyato2018cgans,odena2017conditional}.
On the other hand, our cGAN variant samples a single latent vector $\bz$, which represents class identity and appearance jointly, from $p(\bz|i)=q(\na|i)$, the vMF distribution of class-conditional activations derived in~\cref{subsec:xent}.

Our method is incorporated with SNGAN~\cite{miyato2018cgans} implemented by~\cite{shmelkov2018good}, 
and compared to the original SNGAN on a class-conditional image generation task using the CIFAR10 dataset~\citep{krizhevsky2009learning}.
The only difference between SNGAN and our variant is that we replace the input Gaussian noise $\bz$ of SNGAN with our vMF noise for label $i$, and the input for conditional batch normalization~\cite{dumoulin2016learned} is still $i$.
The vMF distributions used to sample the latent vectors are derived from the WRN-40-2 network~\citep{zagoruyko2016wide} trained on the same dataset. 
Latent vectors of all methods are $128$-dimensional, and we multiply $10$ to latent vectors sampled from our vMF models to match their norm to that of baseline methods.
We directly follow the evaluation protocol of \cite{shmelkov2018good}. 
The only additional hyperparameter was the concentration parameter $\kappa$ of the vMF distribution, which we set to $5$ based on initial experiments.

\input{tables/cGAN.tex}

The quality of generated images is measured in two different metrics, Inception Score (IS)~\cite{salimans2016improved} and Frechet Inception Distance (FID 5k)~\cite{heusel2017gans}.
IS measures the certainty in class prediction along with the diversity between different classes, while FID 5k quantifies the dissimilarity between activations of real and generated images.
Both metrics are based on the ImageNet-pretrained Inception-v3 network~\cite{szegedy2016rethinking}.

The quantitative results in~\Cref{tab:cGAN} show that our model outperforms the baseline, particularly in the IS metric.
We argue that this improvement comes from the advantages of the learned prior $p(\bz|i)=q(\na|i)$ that allow the decoder to utilize the latent space more effectively while considering the affinity between classes and class-specific appearance variations, as discussed in~\cref{subsec:HVAE}.
The qualitative results in~\Cref{fig:gan_img} demonstrate that our method tends to generate images with more diverse instances and backgrounds while keeping their class identity.

%% file: tables/KD.tex
\begin{table}
\caption{ 
    Top-1 test accuracy of the student networks on the CIFAR-100 dataset.
    Results are averaged over $5$ runs.
}
\label{tab:KD}
\centering \begin{tabular}{lS}
\toprule
    Method                              & {Accuracy} \\
\midrule
    Teacher (WRN-40-2)                  & 75.61 \\
\midrule
    Student (WRN-16-2)                  & 73.26 \\
\midrule
    FitNet \cite{romero_fitnets:_2014}  & 73.58 \\
    KD \cite{hinton_distilling_2015}    & 74.92 \\
    AT  \cite{zagoruyko2016paying}      & 74.08 \\
    RKD \cite{park2019relational}       & 73.35 \\
    SP \cite{tung2019similarity}        & 73.83 \\
    VID \cite{ahn2019variational}       & 74.11 \\
    CRD \cite{tian2019contrastive}      & 75.48 \\
\midrule
    CKD (ours)                          & 74.32 \\
    CKD + KD (ours)                     & 75.21 \\
\bottomrule
\end{tabular} 
\end{table}

%% file: tables/KD_resolution.tex
\begin{table*}[t]
\caption{
    Top-1 test accuracy of the student networks on the CIFAR-100 dataset with various degrees of photometric transform and image downsampling. 
    Results are averaged over $5$ runs.
    }
\label{tab:KD_domain}
\centering \begin{tabular}{lSSSSS SSSS}
\toprule
    & \multicolumn{5}{c}{Photometric Transform}
    & \multicolumn{4}{c}{Downsampling} \\ 
    \cmidrule(l{4pt}r{4pt}){2-6} \cmidrule(l{4pt}r{4pt}){7-10}
     & {0.0} & {0.2} & {0.4} & {0.6} & {0.8}
    & {$\times$1.0} & {$\times$0.75} & {$\times$0.5} & {$\times$0.25} \\
\midrule
    Label 
    & 73.26 & 73.29 & 72.97 & 72.39 & 71.53 
    & 73.26 & 70.15 & 63.64 & 49.00 \\
    KD    
    & \BF 74.92 & \BF 74.34 & 71.92 & 65.68 & 51.35
    & \BF 74.92 & 68.52 & 45.84 & 20.27 \\
\midrule
    CKD (ours)   
    & 74.32 & 74.18 & \BF 73.98 & \BF 73.61 & \BF 72.47 
    & 74.32 & \BF 71.24 & \BF 64.55 & \BF 49.72  \\
\bottomrule
\end{tabular} 
\centering \begin{tabular}{lSSS}
\toprule
\midrule
\bottomrule
\end{tabular} 
\end{table*}

%% file: tables/scvae.tex
\begin{table*}[t]
\caption{
    Comparison 
    on the MNIST generative modeling task. 
    Results are averaged over $5$ runs.
    } 
\label{tab:VAE}
\centering
\begin{tabular}{l
    *{2}{S[table-format=-3.1, round-precision=1]}
    *{2}{S[table-format=-2.1, round-precision=1]}
    *{2}{S[table-format=-3.1, round-precision=1]}
    *{2}{S[table-format=-2.1, round-precision=1]}
    }
\toprule
     & \multicolumn{4}{c}{Log-Likelihood} & \multicolumn{4}{c}{ELBO} \\
\cmidrule(l{4pt}r{4pt}){2-5} \cmidrule(l{4pt}r{4pt}){6-9}

Dimension of $\bz$ 
    & {3} & {5} & {10} & {20} 
    & {3} & {5} & {10} & {20} \\
\midrule

HVAE 
    & -122.0162645 & -107.397 & -93.58822 & -90.88662
    & -124.0400988 & -111.283 & -98.01676 & -96.3212 \\
HVAE-C 
    & -124.855 & -110.7306 & -93.3101 & -89.69626
    & -127.6128 & -114.3448 & -97.63816 & -95.27338 \\
\midrule

HVAE-L (ours) 
    &  \BF -119.406 & \BF-105.2332 & \BF-90.50004 & \BF-87.85564
    & \BF -122.9596 & \BF-109.1298 & \BF-95.08924 & \BF-93.34436 \\
\bottomrule
\end{tabular} 
\end{table*}

    
    

%% file: tables/cGAN.tex
\begin{table}[t]
\caption{
    Comparison on the CIFAR10 generative modeling task. 
    Results are averaged over $5$ runs.
}
\label{tab:cGAN}
\centering
\begin{tabular}{
    l
    S[table-format=1.3, round-precision=3]
    S[table-format=2.3, round-precision=3]
    }
\toprule 



$\bz$       & {IS ($\uparrow$)} & {FID 5k ($\downarrow$)}   \\
\midrule
Gaussian  & 8.406             & 18.867    \\
vMF       & \BF 8.511         & \BF 18.764  \\
\bottomrule
\end{tabular} 
\end{table}

%% file: _5_discussion.tex
\section{Discussion}
\label{sec:discussion}

Our core contribution is a simple generative model of activations that forms a generative-discriminative pair with the given classification network.
We believe our approach provides insight into how even the design of a single layer can impose an inductive bias on a model that guides how and where knowledge is stored. 
We show that it is possible to exploit such structures to efficiently extract useful information from a trained model.
While our derivation in \cref{subsec:xent} is specific to the typical design of using matrix multiplication in the final layer (i.e., $\bl = \bW^\top \ba$), 
it could be extended to analyze other multiplicative interactions~\citep{jayakumar2020multiplicative}, including Mahalanobis metric learning, gating mechanisms, and self-attention.

Our approach may be useful in many other applications.
For example, beyond our mild domain shift setting in~\cref{subsubsec:kd_domain_shift}, CKD may also have benefits in a more severe domain adaptation setting where one wishes to transfer information between different domains.
Also, the density functions $q(\na|i)$ may be used as a decision criterion for anomaly detection, and their relative overlap may help estimate or calibrate the uncertainty of classification networks.
We believe these are all exciting directions for future research.



{\small
\noindent \textbf{Acknowledgements:}
This work was supported by 
Kakao,
the NRF grant, 
and the IITP grant, 
funded by Ministry of Science and ICT, Korea
(No.2019-0-01906 AIGS Program--POSTECH,
 NRF-2021R1A2C3012728--60\%, 
 IITP-2020-0-00842--40\%). 
}